\documentclass[sigconf]{acmart}
\usepackage{times}
\usepackage{url}
\usepackage{latexsym}

\usepackage{graphicx} 
\usepackage{subfigure}
\usepackage{colortbl,booktabs}
\usepackage{multirow}
\usepackage{multicol}
\usepackage{tabularx}
\usepackage{setspace}
\usepackage{mathtools}
\usepackage{amsmath,amsfonts} 
\usepackage{threeparttable}
\usepackage{enumerate}
\usepackage{enumitem}
\usepackage{color}
\usepackage{booktabs}
\usepackage{textcomp}

\AtBeginDocument{%
  \providecommand\BibTeX{{%
    \normalfont B\kern-0.5em{\scshape i\kern-0.25em b}\kern-0.8em\TeX}}}

\setcopyright{acmcopyright}
\copyrightyear{2018}
\acmYear{2018}
\acmDOI{10.1145/1122445.1122456}

\acmConference[Woodstock '18]{Woodstock '18: ACM Symposium on Neural
  Gaze Detection}{June 03--05, 2018}{Woodstock, NY}
\acmBooktitle{Woodstock '18: ACM Symposium on Neural Gaze Detection,
  June 03--05, 2018, Woodstock, NY}
\acmPrice{15.00}
\acmISBN{978-1-4503-XXXX-X/18/06}

\begin{document}

\title{CalibreNet: Calibration Networks for Multilingual Sequence Labeling}




\author{Shining Liang}
\authornotemark[1]
\email{liangsn17@mails.jlu.edu.cn}
\affiliation{%
  \institution{College of Computer Science and Technology, Jilin University}
}

\author{Linjun Shou}
\email{lisho@microsoft.com}
\affiliation{%
  \institution{STCA NLP Group, Microsoft}
}

\author{Jian Pei}
\email{jpei@cs.sfu.ca}
\affiliation{%
  \institution{School of Computing Science, Simon Fraser University}
}

\author{Ming Gong}
\email{migon@microsoft.com}
\affiliation{%
  \institution{STCA NLP Group, Microsoft}
}

\author{Wanli Zuo}
\authornotemark[2] 
\email{zuowl@jlu.edu.cn}
\affiliation{%
  \institution{College of Computer Science and Technology, Jilin University}
}

\author{Daxin Jiang}
\authornotemark[2] 
\email{djiang@microsoft.com}
\affiliation{%
  \institution{STCA NLP Group, Microsoft}
}


\renewcommand{\shortauthors}{Shining and Linjun, et al.}

\begin{abstract}

\footnotetext[1]{Work done during the first author's internship at Microsoft STCA.}
\footnotetext[2]{Daxin Jiang and Wanli Zuo are the corresponding authors.}

Lack of training data in low-resource languages presents huge challenges to sequence labeling tasks such as named entity recognition (NER) and machine reading comprehension (MRC).
One major obstacle is the errors on the boundary of predicted answers.
To tackle this problem, we propose CalibreNet, which predicts answers in two steps. In the first step, any existing sequence labeling method can be adopted as a base model to generate an initial answer. In the second step, CalibreNet refines the boundary of the initial answer. To tackle the challenge of lack of training data in low-resource languages, we dedicatedly develop a novel unsupervised phrase boundary recovery pre-training task to enhance the multilingual boundary detection capability of CalibreNet. Experiments on two cross-lingual benchmark datasets show that the proposed approach achieves SOTA results on zero-shot cross-lingual NER and MRC tasks.
\end{abstract}


\begin{sloppy}



\maketitle
\section{Introduction}
\label{intro}
Sequence labeling tasks are essential in web mining, such as named entity recognition (NER)~\cite{whitelaw2008web, li2012twiner, hosseini2019implicit}, event extraction~\cite{zhao2017constructing, deng2020meta}, and relation identification~\cite{lockard2019openceres}. For example, the NER models assign the predefined labels to tag tokens in the input sequences to indicate both the entity boundaries and types~\cite{reimers2017optimal}. In some web services, such as question answering, sequence labeling also plays a critical role, where it reads a passage in a Web page as the context and answers a given question by extracting a text span inside the given passage. This process is often called machine reading comprehension (MRC). MRC is also regarded as a sequence labeling task~\cite{li2016dataset, li2019entity, DBLP:conf/acl/LiFMHWL20}, since it predicts whether each token is the start, end, or none for the answer span.

There is a rich literature for sequence labeling. Classical methods include Hidden Markov models (HMMs)~\cite{zhou2002named}, maximum entropy Markov models (MEMMs)~\cite{mccallum2000maximum}, and conditional random field (CRF)~\cite{lafferty2001conditional}. Recently, combining neural networks as the representation layer with CRF models has further boosted the state-of-the-art performance~\cite{jiang2019improved, baevski2019cloze}. However, such statistical models require large amounts of training data. Consequently, they only show good performance in languages with rich training data, such as English. Sequence labeling on low-resource languages is still very challenging, mainly due to very limited training data available.

To tackle the challenge of sequence labeling in low-resource languages, some early works transfer the knowledge from rich-source languages to low-resource ones by information alignment through manually built bilingual parallel corpora~\cite{wang2014cross},  or language-independent features~\cite{zirikly2015cross,tsai2016cross}. In recent years, multilingual pre-trained language models, such as Unicoder~\cite{huang2019unicoder}, mBERT~\cite{pires2019multilingual}, and XLM-Roberta~\cite{conneau2019unsupervised} (XLM-R), are developed for model transferring. For example, Wu \textit{et al.} \cite{wu2020enhanced} fine-tune mBERT on a pseudo training set by a meta-learning method. To better leverage the unlabeled data in the target language, a teacher-student framework is proposed~\cite{wu2020single} to distill knowledge from weighted teacher models. Inspired by back translation in neural machine translation (NMT), DualBERT~\cite{cui2019cross} is developed to learn source language and target language features simultaneously. Although these multilingual sequence labeling models can effectively locate target spans, they often fail to give the precise boundaries of the spans in the target languages. 

We conduct an empirical study to quantitatively assess the challenge. In Figure~\ref{fig_example} (a), we categorize the mismatches between the predicted span and the ground truth span into four types: (1) the predicted answer is a {\em super span} of the ground truth; (2) the predicted answer is {\em sub span} of the ground truth; (3) the predicted answer both miss some terms in the ground truth and add extra terms not in the ground truth (i.e., \emph{drifted span}), and  (4) the predicted answer is adjacent to the ground truth but contains no common sub-span with it (i.e., \emph{adjacent span}).  We further show in Table~\ref{tab_ner_sts} the statistics of the error cases in the cross-lingual NER task using the XLM-R model, where the boundary errors, including \emph{super span}, \emph{sub span}, \emph{drifted span}, and \emph{adjacent span}, contribute to a large portion of all error cases as shown in the last column. The other errors cases are mainly entity type detection errors. This observation motivates us to tackle the bottleneck of boundary detection in sequence labeling models.

\begin{figure}[!t]
    \centering
    \includegraphics[trim={0.2cm 0.8cm 0cm 0cm},clip,scale=0.28]{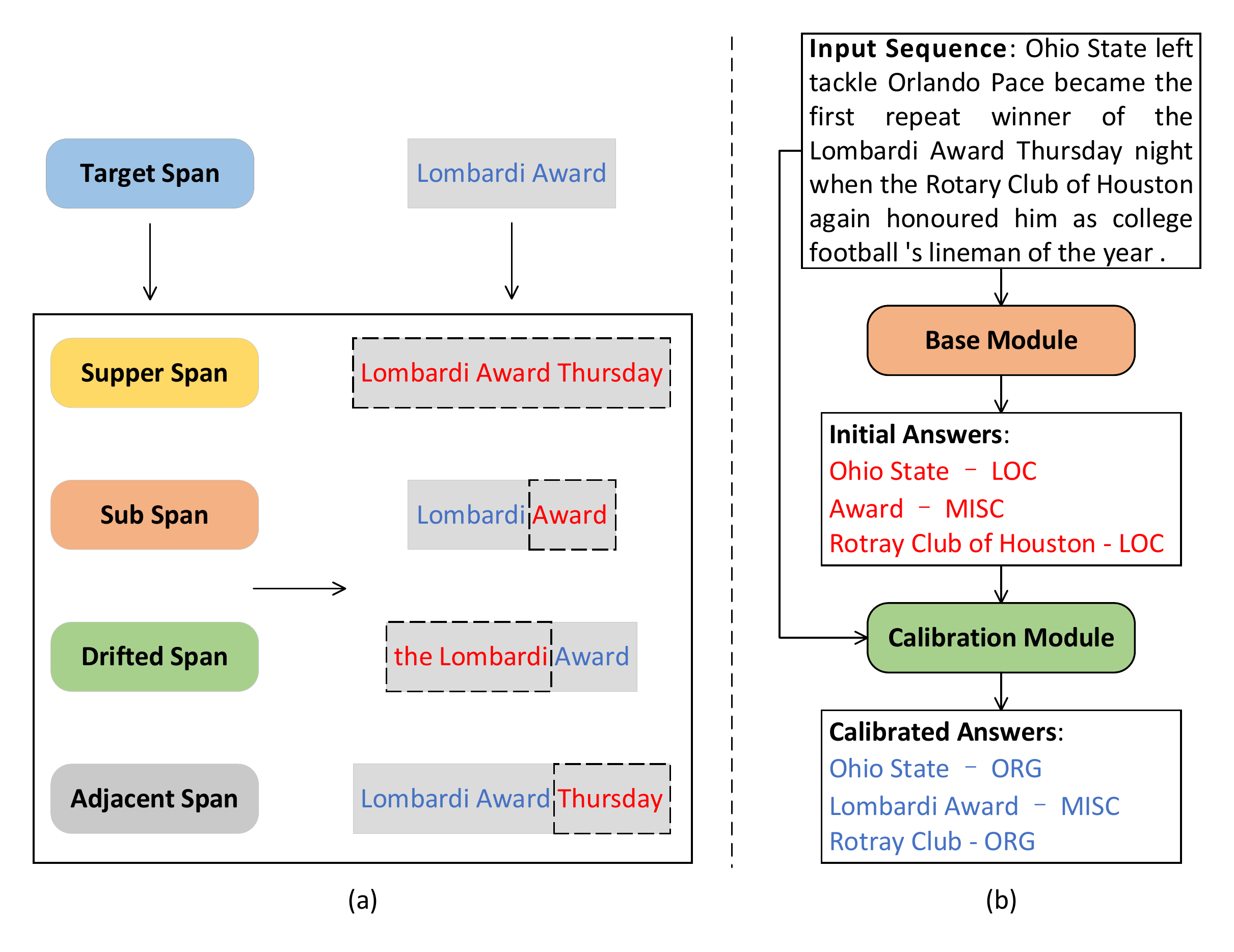}
    \caption{(a) Span boundary overlap phenomenon in cross-lingual NER. Target span is blue; various outputs spans are red and marked with dashed boxes. (b) An overview of our approach. The base module first predicts entities as ``initial answers''. Then the calibration module further refines the initial answers to output the final calibrated answers. }
    \label{fig_example}
\end{figure}

\begin{table}[t]
    \caption{The statistics of the error cases in the XLM-R$_{base}$ baseline on the XGLUE-NER~\cite{Liang2020XGLUEAN} dataset. Detailed definitions of each error category are shown in Figure~\ref{fig_example}(a).} \label{tab_ner_sts}
    \tiny
    \begin{tabular}{l|cc|cccc|c}
    \toprule
    \textbf{} & \textbf{\#Test} & \textbf{\#Error} & \textbf{\#Super} & \textbf{\#Sub} & \textbf{\#Drifted} & \textbf{\#Adjacent} & \textbf{Boundary error (\%)} \\
    \midrule
    en & 6,392  & 566   & 106 & 64  & 1 & 137 & 54.4 \\
    es & 4,054  & 955   & 93  & 246 & 0 & 295 & 66.4 \\
    de & 5,390  & 1,648 & 201 & 150 & 4 & 884 & 75.2 \\
    nl & 6,884  & 949   & 148 & 96  & 0 & 42  & 30.1 \\
    \bottomrule
    \end{tabular}
    \vspace{-5pt}
\end{table}

Accurately detecting answer boundaries becomes a bottleneck in sequence labeling.  To tackle the challenge, in this paper, we propose a separate model for boundary calibration based on the output of a base model. Intuitively, the base model captures the global context of the whole input sequence and roughly locates the region for answers. Then, the calibration model conducts finer search within the detected region and the neighborhood, and focuses on the local context to refine the boundary. This is analogous to the human perception and cognition process~\cite{evans2008dual}, which first locates the target, sets up the local context, and finally zooms into details.  Our design is novel for sequence labeling, and is orthogonal and complements to all existing approaches.

Using a second model to focus on detecting answer boundaries accurately is an intuitive and nice idea.  However, how to construct high-quality training data for the calibration model remains challenging. One straightforward method is to transform the original training data of sequence labeling task into a new training set for calibration model. However, the data collected in this way is still quite limited, especially for low-resource languages. To address this challenge, we strategically propose a novel phrase boundary recovery (PBR) task to pre-train the model on large-scale augmented datasets synthesized from Wikipedia documents in multiple languages. The new pre-training approach dramatically improves the capability of the calibration module to determine answer boundaries accurately.


Our approach is shown in Figure~\ref{fig_example}(b). CalibreNet consists of two modules, a base module and a calibration module. The base module can take any model of sequence labeling. The predicted answers by the base module are combined with the input sequence to form the input to the calibration module. The calibration module considers both the initial results by the base module and the whole passage to refine the span boundaries. In particular, the calibration module is pre-trained with the PBR task on large-scale multilingual synthesized data from Wikipedia-derived corpus.

We make the following technical contributions in this paper. First, we propose the CalibreNet framework for the task of cross-lingual sequence labeling to improve the accuracy of labeled answers. Second, we propose a novel phrase boundary recovery task and a weakly supervised pre-training method using Wikipedia data. This approach effectively enhances the model sensitivity to phrase boundaries. Last but not least, we conduct extensive experiments on zero-shot cross-lingual NER and improve the SOTA results. In addition, the experiments on the MRC tasks also show consistent improvement over strong baseline methods.

The rest of the paper is organized as follows. We first review the related work in Section~\ref{sec:rela}. We then present our approach in Section~\ref{sec:method}. We report the extensive experimental results in Sections~\ref{sec:experiment}.  We conduct further analysis in Section~\ref{sec:ablation}, and conclude the paper in Section~\ref{sec:conclusion}.

\section{Related Work}\label{sec:rela}

To tackle the challenge of having only very limited training data in low-source languages for sequence labeling tasks, two major approaches are explored.  The first approach is to transfer knowledge from a source language with rich labeled data to a target language with little or even no labeled data. This is called cross-lingual sequence labeling. The other approach is to look for additional data sources and conduct weakly supervised learning. 

In this paper, we focus on two multilingual sequence labeling tasks, NER and MRC.


The previous methods for cross-lingual named entity recognition (NER) can be divided into data transfer methods and model transfer methods, according to the knowledge transfer mechanisms. Data transfer methods generate annotations in target languages that are used for model training. For example, \citeauthor{wang2014cross} employ bilingual parallel corpora and transfer labels through word alignment. Since parallel data may not be available, some other methods apply machine translation as a substitution for parallel data. For example, phrase alignment~\cite{mayhew2017cheap} or word alignment~\cite{xie2018neural} can be used to transfer labels. However, the performance of these methods is limited by the translation quality, especially for low-resource languages. 

The idea of model transfer is to generate language-independent features for NER models. By training models on source language, those methods automatically scale out to other languages through language-independent features. Some representative methods leverage gazetteers~\cite{zirikly2015cross} and Wikifier features~\cite{tsai2016cross}, and align word representations~\cite{ni2017weakly}. Most recent studies~\cite{wu2019beto,moon2019towards} make great improvements by multilingual pre-trained language models. To reduce the dependency on training data in source languages, Wu \textit{et al.}~\cite{wu2020single} propose a knowledge distillation method, which distills knowledge from a multilingual teacher model to a student model through unlabeled data in target languages. Wang \textit{et al.}~\cite{DBLP:conf/acl/WangJBWHT20} develop another knowledge distillation method to transfer the structural knowledge from several monolingual models into a single multilingual model. Their motivation is to reduce the gap between a universal model for multiple languages with individual monolingual models. Our method also belongs to the model transfer category. We tackle the inaccurate boundaries of predicted entities, an issue overlooked by the previous methods. Therefore, all previous methods can be adopted as the base module in our approach.

Different from previous methods, which transfer knowledge from source language, Cao \textit{et al.}~\cite{cao2019low} propose to generate weakly labeled data from Wikipedia pages in low-resource languages. Our method also uses Wikipedia anchor text as ground truth entities, but we do not directly use them for training a NER model. Instead, we synthesize initial answers and pre-train the model for entity boundary detection.

Machine reading comprehension (MRC) in rich-resource languages, such as English and Chinese, has been intensively studied in the past years. However, there are only few studies on cross-lingual MRC, partially due to the lack of data sets until most recently. Artetxe \textit{et al.}~\cite{artetxe2019cross} extract instances from SQuAD v1.1 and translate them into ten languages in total by experts. \citeauthor{lewis2019mlqa} release MLQA where the parallel  sentences containing the answers are extracted with surrounding text from Wikipedia articles and human experts are employed to translate the questions. The availability of benchmark datasets facilitates the development of cross-lingual MRC methods. Hsu \textit{et al.}~\cite{hsu2019zero} demonstrate the feasibility of applying multilingual pre-trained language models for zero-shot MRC. Cui \textit{et al.}~\cite{cui2019cross} propose a Dual-BERT architecture to encode $\langle$\emph{question, passage}$\rangle$ pairs in source and target languages together and use modified multi-head attention to enhance the information fusion. Similar to the previous NER methods, the existing MRC methods cannot handle the boundaries of answers well. Our approach provides a general framework for sequence labeling tasks, in which MRC fits well through a simple extension to the setting for NER.

\begin{figure*}[!htbp]
    \centering
        \subfigure[CalibreNet Architecture]{
            \label{fig_finetuning}
            \includegraphics[scale=0.30]{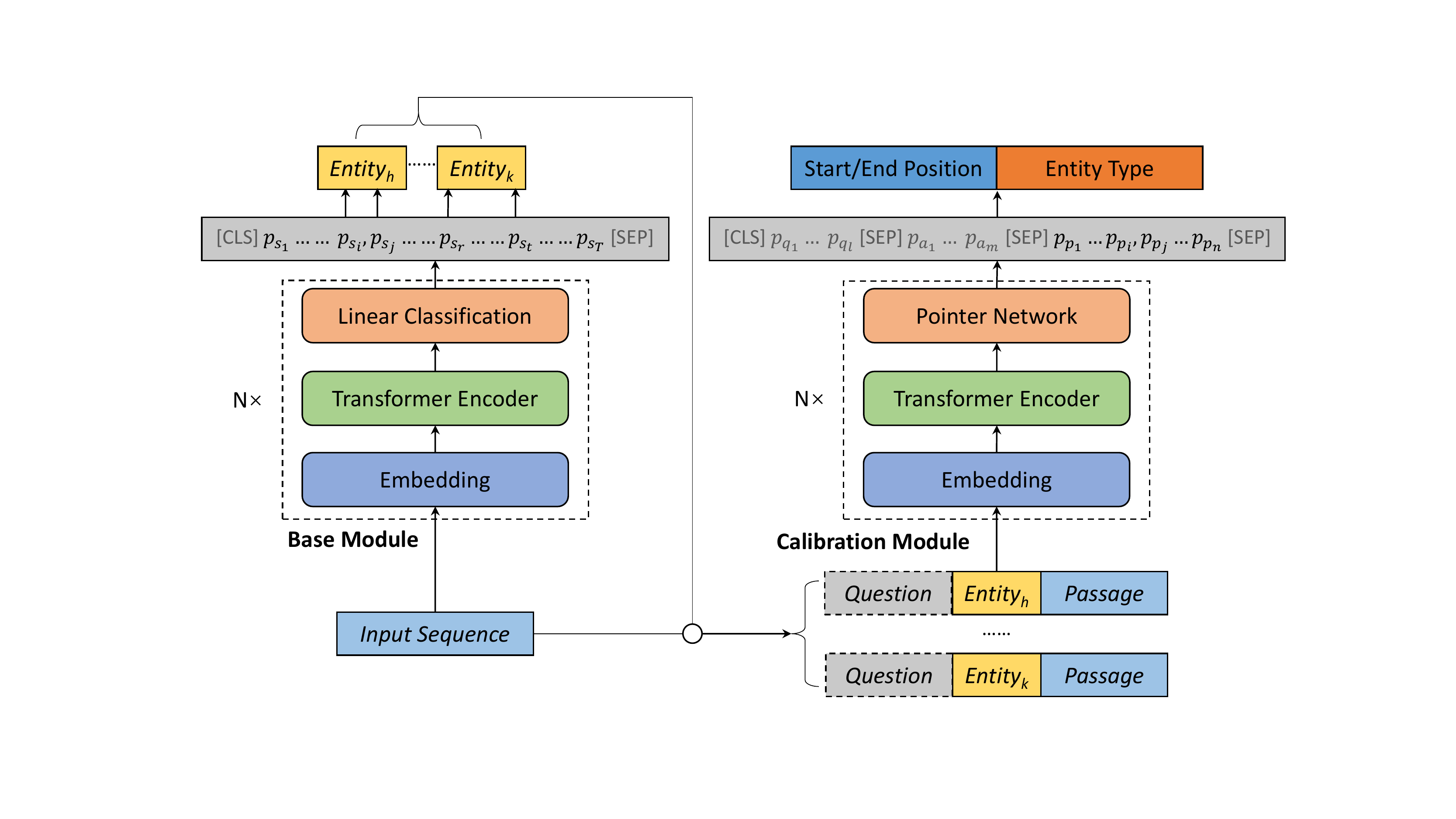}}
        \subfigure[Phrase Boundary Recovery Pre-training]{
            \label{fig_pre-training}
            \includegraphics[trim={6cm 3.7cm 4.5cm 2.5cm},clip,scale=0.32]{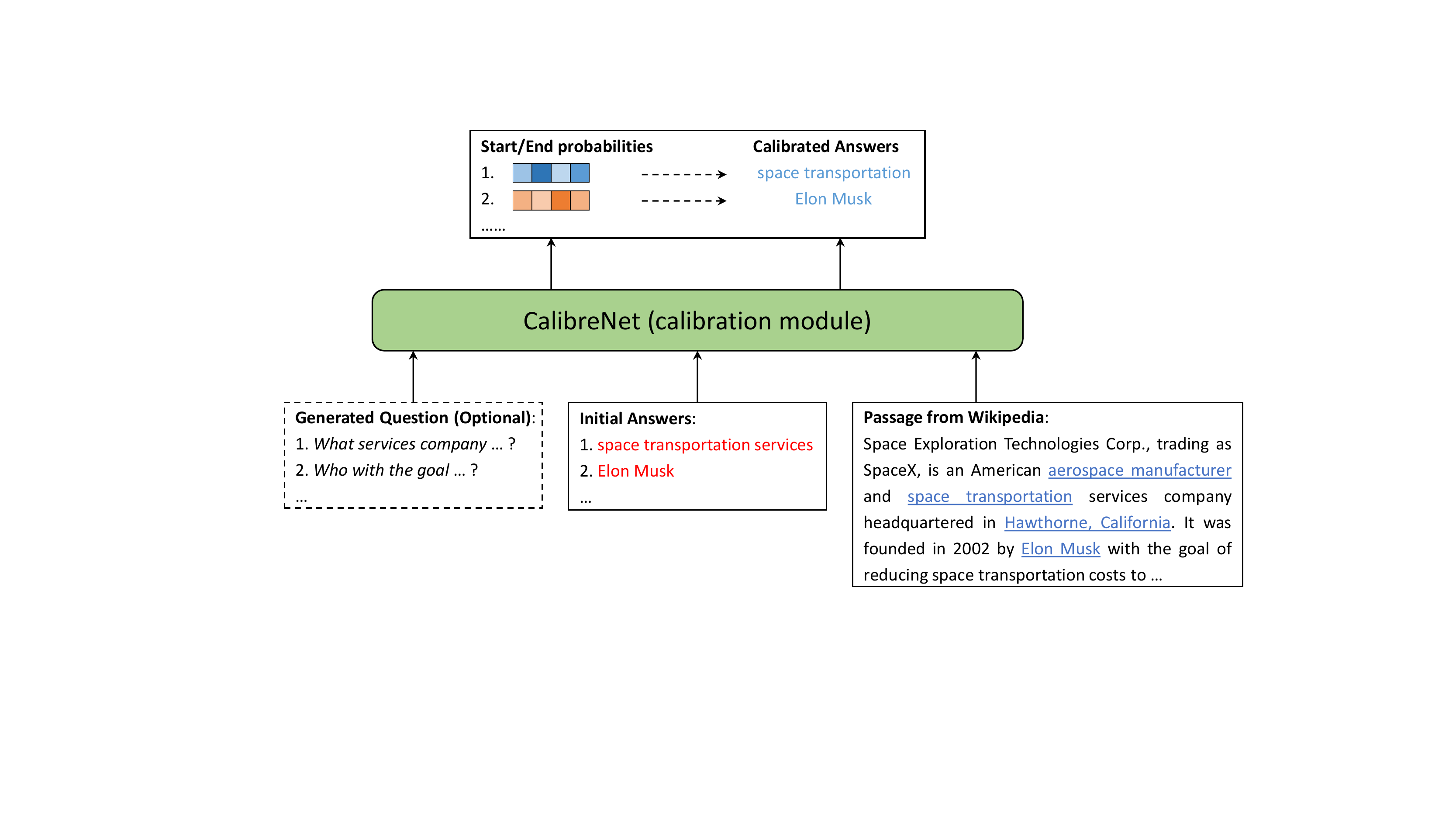}} \\
    \caption{The model architecture of CalibreNet for cross-lingual sequence labeling tasks. The two modules have the same layers in the two dashed boxes but are trained independently.} \label{fig_calinet}
\end{figure*}

\section{CalibreNet}
\label{sec:method}

In this section, we describe CalibreNet for sequence labeling. We first focus on the NER task and elaborate the model architecture (Section~\ref{chp_archi}) and the model pre-training mechanism (Section~\ref{chp_pbr}). Then we extend our approach to tackle the MRC task (Section~\ref{chp_mrc}).

\subsection{CalibreNet Architecture} \label{chp_archi}
The architecture of CalibreNet for sequence labeling is shown in Figure~\ref{fig_finetuning}. In general, the architecture can use any sequence labeling model as the base module. We take the transformer based NER model~\cite{wu2020single} as an example for the base module in the rest of the paper to make the presentation clear. In such a base module, the output layer is a linear classification layer for the NER task. The calibration module is a secondary decoder on top of the base module, which outputs the start/end positions as well as the type of the detected entity.

\begin{figure}[H]
    \centering
    \includegraphics[trim={0cm 0.4cm 0cm 0.65cm},clip,scale=0.3]{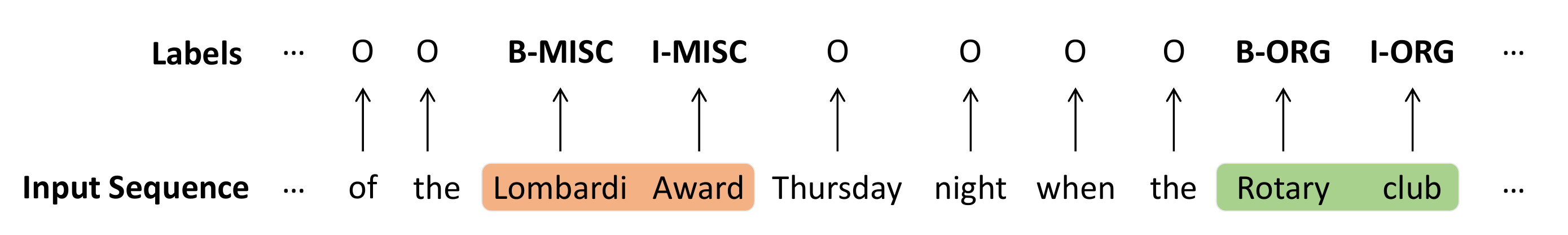}
    \caption{The input sequence and labels of a NER example.}
    \label{fig_input}
    \vspace{-5pt}
\end{figure}

\subsubsection{Base Module} 

The input sequence of the base module is $X=\{x_k\}_{k=1}^n$ with $n$ tokens and the corresponding label sequence is $Y=\{y_k\}_{k=1}^n$ as shown in Figure~\ref{fig_input}.  We feed $X$ into the encoder $\mathcal{E}_B$ to obtain the hidden vectors $\boldsymbol{H}$, that is,
\begin{align}
    \boldsymbol{H} = \mathcal{E}_B(X)
\end{align}
where $\mathcal{E}_B$ is the last layer of a transformer based multilingual pre-trained language model, such as mBERT~\cite{pires2019multilingual} or XLM-R~\cite{conneau2019unsupervised}. $\boldsymbol{H} \in \mathbb{R}^{n\times d}$ is the representation matrix, where a row is the feature for the $k$-th token $x_k$, and $d$ is the vector dimension of the encoder. The hidden vector $\boldsymbol{H}_k$ for each token $x_k$ is then fed into a linear classification layer followed by the \emph{softmax} function to output the probability distribution of entity labels. The loss function $\mathcal{L}_{B}$ of the base module is defined as follows:
\begin{align}
    \boldsymbol{p}_k &= \text{\emph{softmax}}(\boldsymbol{W}_B \cdot (\boldsymbol{H}_k)^T + \boldsymbol{b}) \\
    \mathcal{L}_{B} &= \sum_{k=1}^{n}-y_k \log\boldsymbol{p}_k
\end{align} 
where $\boldsymbol{p}_k \in \mathbb{R}^{1 \times \left| C \right|}$, $C$ is the entity labeling set under the BIO annotation schema~\cite{ramshaw1999text}, and $\boldsymbol{W}_B \in \mathbb{R}^{\left| C \right| \times d}$ and $\boldsymbol{b}$ are the output matrix and bias, respectively.

\subsubsection{Calibration Module} 

Given an initial answer $A=\{a_j\}_{j=1}^m$ (predicted entity) and the input text sequence $X=\{x_k\}_{k=1}^n$ from the base module, we need to find the golden answer $y$ (ground truth entity) in $X$. The input to the calibration module is concatenated into a sequence as:
\begin{align}
    \bar{X} = \left\{\left[\rm CLS\right] \{a_j\}_{j=1}^m \left[\rm SEP\right] \{x_k\}_{k=1}^n [SEP] \right\}
\end{align}
The calibration module shares the same encoder structure with the base module. The sequence representations $\bar{\boldsymbol{H}} \in \mathbb{R}^{L\times d}$ can be obtained from the encoder $\mathcal{E}_C$, where $L$ is the maximum sequence length. $\bar{\boldsymbol{H}}$ is fed into a pointer network~\cite{seo2016bidirectional} to locate the answer position. We train the model and take the start and end positions of $y$ as the optimization objective, that is,
\begin{align}
    \boldsymbol{p}^{start} &= \text{\emph{softmax}}(\boldsymbol{W}_{start} \cdot (\bar{\boldsymbol{H}})^T + \boldsymbol{b}_{start}) \\
    \boldsymbol{p}^{end} &= \text{\emph{softmax}}(\boldsymbol{W}_{end} \cdot (\bar{\boldsymbol{H}})^T + \boldsymbol{b}_{end}) \\
    \mathcal{L}_{start,end} &= -\frac{1}{2}(y_{start} \log\boldsymbol{p}^{start} + y_{end} \log\boldsymbol{p}^{end})
\end{align}
where $\boldsymbol{p}_{start} \in \mathbb{R}^{1 \times L}$, $\boldsymbol{p}_{end} \in \mathbb{R}^{1 \times L}$ are probability distributions for start and end positions, and $\boldsymbol{W}_{start} \in \mathbb{R}^{1 \times d}$, $\boldsymbol{W}_{end} \in \mathbb{R}^{1 \times d}$ are trainable matrices for start and end positions, respectively, $\boldsymbol{b}_{start}$ and $\boldsymbol{b}_{end}$ are the biases, $\mathcal{L}_{start,end}$ is the answer position loss function.

\begin{table}[t]
    \caption{Upper-bound analysis when the entity boundary or entity type is assumed correct (i.e., oracle result), conducted with XLM-R$_{base}$ model on XGLUE-NER dataset. Metrics are presented in F1 score with \% omitted.} \label{tab_upper}
    \resizebox{1.0\linewidth}{!}{
    \small
    \begin{tabular}{l|ccccc}
    \toprule
    & \textbf{test-en} & \textbf{test-es} & \textbf{test-de} & \textbf{test-nl} & \textbf{Avg.} \\
    \midrule
    XLM-R$_{base}$          & 91.2 & 76.6 & 69.5 & 78.7 & 79.0 \\
    assume boundary correct & 92.9 (+1.7) & 82.7 (+6.1) & 76.3 (+6.8)  & 80.9 (+2.2) & 83.2 (+4.2) \\
    assume type correct     & 95.8 (+4.6) & 85.4 (+8.8) & 82.5 (+13.0) & 92.8 (14.1) & 89.1 (+10.1)\\
    \bottomrule
    \end{tabular}
    }
    \vspace{-8pt}
\end{table}

In addition to span positions, it is also critical to predict entity types in the NER task, such as PER, ORG, and LOC. Both the entity boundaries and entity types affect NER performance.  For example, Table~\ref{tab_upper} shows the upper-bound analysis when the entity boundaries or entity types of predicted results from the XLM-R$_{base}$ model are assumed correct separately. Therefore, our model should also output the answer types according to the predicted answer spans. CalibreNet extracts the hidden representations $\bar{\boldsymbol{H}}_{\boldsymbol{p}^{start}_{max}}$ and $\bar{\boldsymbol{H}}_{\boldsymbol{p}^{end}_{max}}$ of the tokens with the maximum probabilities of start and end position and feeds them to a linear classification layer.
\begin{align}
    \boldsymbol{p}^{cls} &= \text{\emph{softmax}}(\boldsymbol{W}_{cls} \cdot \frac{1}{2}(\bar{\boldsymbol{H}}_{\boldsymbol{p}^{start}_{max}} + \bar{\boldsymbol{H}}_{\boldsymbol{p}^{end}_{max}})^T + \boldsymbol{b}_{cls}) \\
    \mathcal{L}_{cls} &= - y_{cls}\log\boldsymbol{p}^{cls}
\end{align}
where $\boldsymbol{W}_{cls} \in \mathbb{R}^{\left| C \right| \times d}$ is the weight matrix, $y_{cls}$ is the entity type of the golden answer, $\mathcal{L}_{cls}$ is the entity type classification loss.

One challenge in the calibration module happens when there are multiple mentions of the same entity in the input sequence. The calibration module may be confused about which mention to calibrate. Thus, we propose to introduce the position information of the detected initial answer to distinguish the mention under calibration from the others.  Specifically, the index set for the initial answer in $\bar{X}$ is defined as follows.
\begin{equation}
    \boldsymbol{I} =
    \left\{ \begin{aligned}
        & 0    && {i \in [1, 1+m]} \\
        & \chi(\bar{x}_i) && {i \in (1+m ,L]} \\
    \end{aligned}
    \right.
\end{equation}
where $\chi(\bar{x}_i) \in \{0, 1\}$ is an indicator function to indicate whether the $i$-th token in the input sequence belongs to the currently processed initial answer in the passage. We map $\boldsymbol{I}$ to a trainable index embedding $\boldsymbol{E}_I$, named \emph{initial-index embedding}, and add it to the embedding layers, including the word embedding $\boldsymbol{E}_W$ and the positional embedding $\boldsymbol{E}_P$, which explicitly indicates the position of the current initial answer in the context. 

Besides introducing the position information of the initial answer, we further apply a \emph{span matching} method similar to~\cite{DBLP:conf/acl/LiFMHWL20} at the output layer, which aims to match a predicted start
index with its corresponding end index of the golden entity span. Concretely, the feature vector of each token in $\bar{X}$ is matched with the others by concatenation to represent the start and end indexes separately. Next, a binary classification layer is used to predict whether the given start index and end indexes belong to one golden entity span as follows.
\begin{align}
    \boldsymbol{p}^{idx}_{i_{start},j_{end}} &= {sigmoid}(\boldsymbol{W}_{idx} \cdot (\bar{\boldsymbol{H}}_{i_{start}} \| \bar{\boldsymbol{H}}_{j_{end}})^T + \boldsymbol{b}_{idx}) \\
    \mathcal{L}_{span} &= -y_{start,end}\log\boldsymbol{p}^{idx}
\end{align}
where $\boldsymbol{p}^{idx}_{i_{start},j_{end}}$ indicates that index $i$ and index $j$ should be matched since they belong to the same golden entity, $\boldsymbol{W}_{idx} \in \mathbb{R}^{1 \times 2d}$ is the weight matrix, $\|$ indicates the concatenation function, $y_{start, end}$ denotes the golden labels of whether the start index $i$ should be matched with the end index $j$ since they belong to the same golden answer, and $\mathcal{L}_{span}$ is the start-end index matching loss.  

Finally, we train the three objective functions jointly by
\begin{align}
    \mathcal{L_D} = \alpha \mathcal{L}_{start,end} + \beta \mathcal{L}_{cls} + \gamma \mathcal{L}_{span}
\end{align}
where $\alpha, \beta, \gamma \in [0, 1]$ are the trade-off hyper-parameters for the final loss.

In terms of training data for the calibration module, we leverage the training data and predicted output of base module to derive the data. To be specific, the text sequence $X=\{x_k\}_{k=1}^n$ and the initial answer $A=\{a_j\}_{j=1}^m$ predicted by the base module are taken as input and the golden label $Y=\{y_k\}_{k=1}^n$ of $X$ as output. When there are multiple initial answers in a single input sequence, we create multiple training instances for each initial answer. In the meanwhile, usually, the base module based on large-scale pre-trained multilingual language models already achieves decent performance, the fine-tuning data created above may be imbalanced, that is, most of the predicted entities are correct. To help the model be even better trained on error cases, we artificially introduce more garble into the output of the base module using the method similar to that in Section~\ref{chp_pbr}. Those augmented cases are appended to the training data set created above.

\subsection{Pre-training Strategy} \label{chp_pbr}

To enhance the capability of the calibration module for precise boundary detection, we design a novel pre-training method, called \emph{Phrase Boundary Recovery} (PBR). The idea is to synthesize a large number of noisy entity spans by garbling the true entity phrases. As shown in Figure~\ref{fig_pre-training}, the garbled entity spans are fed into the calibration module together with the original input sequence where the spans reside, and the training target is to recover the true entity spans. Such synthetic data actually simulates the output of the base module. Therefore, if the calibration module gets sufficient pre-training by this task, it can improve the performance of the calibration module, and thus the whole NER task. As the model needs to recover the start and end position of ground truths, we formulate a MRC loss function as the training objective of PBR. Besides, PBR also integrates masked language modeling (MLM)~\cite{devlin2019bert} training objective in a multi-task setting. 
In the following, we introduce how to collect the data sources with passages and correctly labeled entity spans, how to generate the noisy entity spans, and how to develop an effective pre-training strategy to achieve good performance on multiple languages with a single multilingual model.

\begin{figure}[t!]
    \centering
    \includegraphics[width=1.0\linewidth]{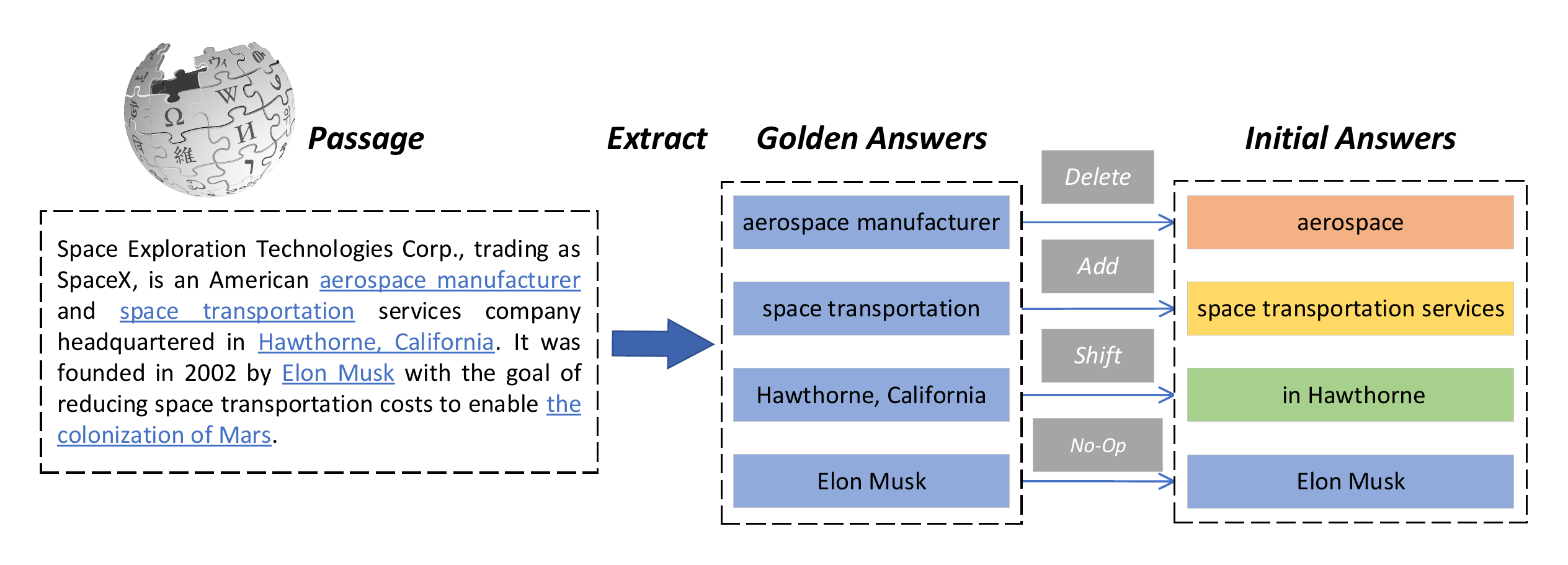}
    \caption{An example of extracting golden answers and generating initial answers.}
    \label{fig_pbr}
    \vspace{-5pt}
\end{figure}

\subsubsection{Data Collection} 
For the NER task, the training data usually comprises passages as the input sequences and the corresponding labeled entities as the training target within the passages. We consider constructing the training data from Wikipedia, which contains enormous multilingual web pages. We randomly extract passages with 50 to 250 tokens from Wikipedia dumps\footnote{The timestamp of all the dumps in different languages we used is 20200301.}, with the section titles removed. To derive labeled entities, we leverage the anchor links in the Wikipedia pages, that is, the anchor texts marked by the hyperlinks are considered as the ground truth entities. We found that the majority of anchor texts are entities and noun phrases, which meet the requirement of NER and MRC well. Based on our empirical study, several intuitive rules are applied to the raw data to improve data quality. Specifically, we first iterate over passages and discard those with only one anchor text. Then we only keep the anchor texts up to eight tokens in the training data. 

In this paper, we choose Wikipedia pages in English (en), Spanish (es), and German (de) to pre-train the calibration module and demonstrate the effectiveness of our approach. In practice, we can choose the languages in the pre-training stage according to the target languages as well as the time and cost constraints. The statistics of our pre-training data are shown in Table~\ref{tab_pbr_sts}.

\begin{table}
\small
    \caption{Statistics of the pre-training data.} \label{tab_pbr_sts}
    \begin{center}
        \begin{tabular}{lccc}
            \toprule
            & \textbf{en} & \textbf{es} & \textbf{de} \\
            \midrule
            \# passages             & 98k   & 44k  & 93k   \\
            \# answers              & 450k  & 456k & 457k  \\
            Avg. answer tokens      & 1.99  & 1.86 & 1.64  \\
            Avg. answer\,/\,passage & 4.59 & 10.34 & 4.93  \\
            \bottomrule
        \end{tabular}
    \end{center}
    \vspace{-10pt}
\end{table}


\subsubsection{Initial answer generation}
Since the function of the calibration module is to recover precise entity boundaries from the initial and possibly inaccurately predicted answers by the base module, we still need to synthesize initial answers to simulate the input to the calibration module when it is combined with the base module in practice. Our method is to garble the known ground truth answers with some operations, following a sampling strategy according to the statistics of errors given in Table~\ref{tab_ner_sts}. Specifically, we build an empirical distribution $p(op, n_{op})$ from the statistics for Table~\ref{tab_ner_sts},  where $op$ is one of the following operations, and $n_{op}$ is the number of tokens to be involved in the operation. Given a ground truth answer, we sample from $p(op, n_{op})$ and conduct one of the following operations according to the sampled $op$. 
\begin{itemize}
    \item \emph{No-Op}: Directly copy the ground truth answer without any change. This corresponds to the cases when the initial answer by the base module perfectly matches the ground truth.
    
    \item \emph{Add}: $n_{op}$ tokens are randomly attached to the left, right, or both sides of the ground truth answer to derive an initial answer. This corresponds to the {\em super span} error in Figure~\ref{fig_example}.

    \item \emph{Delete}: $n_{op}$ tokens are randomly removed from the left, right, or both sides of the ground truth answer to derive an initial answer. This corresponds to the {\em sub span} error in Figure~\ref{fig_example}.

    \item \emph{Shift}: Keep the length of the answer unchanged, but the start position of the ground truth answer is randomly moved to the left or right by $n_{op}$ tokens to derive an initial answer. When $n_{op}$ is smaller than the length of the ground truth answer, this corresponds to the {\em drifted span} error in Figure~\ref{fig_example}. If $n_{op}$ equals to the length of the ground truth answer, it becomes the case of {\em adjacent span}. 
\end{itemize}

\subsubsection{Continual Multilingual Learning}
To pre-train the CalibreNet in multiple languages, one straightforward method is to train in each language corpus sequentially. However, this method may lead to the catastrophic forgetting phenomenon. Inspired by continual multi-task learning in ERNIE~\cite{sun2020ernie}, we propose \emph{continual multilingual learning} for pre-training the calibration module. We divide the training into several stages. In each stage, the model is continuously trained on the training data of a new language as well as a randomly sampled subset of the training data in the previous stages. In this way, the model can gradually acquire the knowledge in each language without catastrophic forgetting of the previously trained languages. Detailed experiments will be discussed in Section~\ref{effect_pretrain}.

\subsection{Applying CalibreNet to MRC} \label{chp_mrc}

In the previous sections, we use NER as an example to describe the CalibreNet approach. In fact, our approach is generally applicable to all sequence labeling tasks. Let us show how to apply our method to the MRC task.

Different from the NER task, where the input is a single sequence of tokens, the input to the MRC task consists of two sequences, a question and a passage. The passage is where the answers to be labeled, therefore, it is similar to the input to the NER task. However, the questions cannot be ignored when training the MRC model since the questions have a strong semantic correlation with the answers to be labeled. Therefore, when we construct pre-training data using Wikipedia as introduced in Section~\ref{chp_pbr}, we also need to consider how to generate questions for the MRC task.

In this paper, we use a simple question generation method similar to~\cite{DBLP:conf/acl/FabbriNWNX20}. Suppose the sentence containing the ground truth answer can be split in the form of [\emph{Fragment A}] [\emph{Answer}] [\emph{Fragment B}], we generate questions according to the template ``Wh+B+A+?'' by replacing the answer with a Wh-component (e.g., what, who, where), which depends on the type of the answer. Specifically, we first locate the ground truth answer (i.e., an anchor text) in the sentence and extract the clause containing it. If the clause is longer than ten tokens, we clip a ten tokens context window from the sentence. Second, we use the spaCy\footnote{https://spacy.io/} for entity type recognition of the ground truth answer and accordingly select the Wh-component. We then follow the method in Section~\ref{chp_pbr} to generate an initial answer, and finally obtain a $\langle$\emph{question, initial answer, passage}$\rangle$ tuple as a pre-training example.

We make up the input sequence to the calibration module in fine-tuning step by adding the question $Q=\{q_i\}_{i=1}^l$ at the beginning:
\begin{align}
    \bar{X} = \left\{\left[\rm CLS\right] \{q_i\}_{i=1}^l \left[\rm SEP\right] \{a_i\}_{j=1}^m \left[\rm SEP\right] \{p_k\}_{k=1}^n [SEP] \right\}
\end{align}
We find that the components of initial-index embedding and span matching in the loss function defined in Section~\ref{chp_archi} are effective for CalibreNet. Since the type of answers in MRC is strongly correlated with the question, the extra loss from entity type has little impact on the results. Therefore, we drop the loss from $\mathcal{L}_{cls}$ and the loss function for MRC becomes $\mathcal{L}_D = \frac{1}{2}(\mathcal{L}_{start,end} + \mathcal{L}_{span})$.

\section{Experiments}\label{sec:experiment}

In this section, we report a comprehensive empirical evaluation of our proposed CalibreNet method.

\subsection{Datasets and Evaluation Metrics}

We conduct experiments and analyses on two cross-lingual sequence labeling datasets: XGLUE-NER and MLQA. \textbf{XGLUE-NER} is selected from CoNLL-2002 NER~\cite{DBLP:conf/conll/Sang02} and CoNLL-2003 NER~\cite{DBLP:conf/conll/SangM03}. It includes English, German, Spanish, and Dutch. There are four types of named entities: PER, LOC, ORG, and MISC. \textbf{MLQA}~\cite{lewis2019mlqa} is an extractive multilingual question answering benchmark of which the passages are aligned on Wikipedia articles. Here we select the cross-lingual transfer task (XLT) except zh (Chinese). 

For both datasets, we follow the zero-shot setting, which means we fine-tune the model on English train set, use the English dev set to select hyperparameters and then infer on every target language test sets. Table~\ref{tab_datasts} provides the detailed statistics of the XGLUE-NER and MLQA datasets.

\begin{table}[t]
    \caption{Experiment Dataset Statistics.} \label{tab_datasts}
    \begin{center}
    \tiny
    \begin{tabular}{l|ccccccccc}
    \toprule
    \multirow{2}{*}{\textbf{Dataset}} & \multirow{2}{*}{\textbf{Train}} &\multirow{2}{*}{\textbf{Dev}} & \multicolumn{7}{c}{\textbf{Test}} \\
    \cmidrule{4-10}
    & & & en & es & de & ar & hi & vi & nl \\ 
    \midrule
    XGLUE-NER   & 14,042 & 3,252 & 3,454  & 1,523 & 3,007 &  -   &  -   &  -   & 5,202 \\
    MLQA        & 87,599 & 1,148 & 11,590 & 5,253 & 4,517 & 5,335 & 4,918 & 5,495 & -    \\
    \bottomrule
    \end{tabular}
    \end{center}
    \vspace{-15pt}
\end{table}

Following the evaluation in XGLUE~\cite{Liang2020XGLUEAN} paper, we use entity-level F1 score for XGLUE-NER. And following the XLM-R paper~\cite{conneau2019unsupervised}, we use span-level F1 and EM (exact match) scores for MLQA. Entity-level F1 score requires the boundary and type of the predicted entity and the ground truth entity to be exactly matched while span-level F1 score can measure the answer overlap between predictions and ground truth. Exact Match score reflects the model capability of matching the answer boundary precisely, which can be regarded as the percentage of cases where the predicted answer matches the ground truth completely.

\begin{table*}[t]
\caption{The overall evaluation results on XGLUE-NER (a) and MLQA (b) datasets. $\P$ indicates statistically significant improvements over baseline methods.}\label{tab_main_compare}
\vspace{-15pt}
\subtable{\small \textbf{(a) Results on XGLUE-NER (F1).}}
{
\small
    \begin{center}
    \begin{tabular}{ll|cccc|c}
    \toprule
    \textbf{Model} & \textbf{Methods} & \textbf{en} & \textbf{es} & \textbf{de} & \textbf{nl} & \textbf{Avg.}\\
    \midrule
    \multirow{5}{*}{mBERT}  & XGLUE-NER$_{\mbox{org}}$  & 90.60 & 75.40 & 69.20 & 77.90 & 78.20 \\
    & XGLUE-NER$_{\mbox{re-imp}}$ & 90.59 & 74.84 & 70.46 & 77.87 & 78.44 \\
    & CalibreNet (XGLUE-NER$_{\mbox{re-imp}}$) & 91.25 & 76.10 & 73.06 & 79.70 & \textbf{80.03}$\P$ \\
    & Single-source KD & - & 76.94 & 73.22 & 80.89 & 77.02 \\
    & CalibreNet (Single-source KD) & - & 77.36 & 75.47 & 81.61 & \textbf{78.15}$\P$ \\
    \midrule
    \multirow{3}{*}{XLM-R$_{base}$} & XGLUE-NER$_{\mbox{org}}$ & 90.90 & 75.20 & 70.40 & 79.50 & 79.00 \\
    & XGLUE-NER$_{\mbox{re-imp}}$   & 91.20 & 76.60 & 69.50 & 78.69 & 79.00 \\
    & CalibreNet (XGLUE-NER$_{\mbox{re-imp}}$) & 91.00 & 78.04 & 74.10 & 80.55 & \textbf{80.92}$\P$ \\
    \bottomrule
    \end{tabular}
    \end{center}
}
    
\subtable{\small \textbf{(b) Results on MLQA (F1/EM).}}
{
\small
    \begin{center}
    \begin{tabular}{ll|cccccc|c}
    \toprule
    \textbf{Model} & \textbf{Methods} & \textbf{en} & \textbf{es} & \textbf{de} & \textbf{ar} & \textbf{hi} & \textbf{vi} &  \textbf{Avg.}\\
    \midrule
    \multirow{2}{*}{mBERT} & MLQA$_{\mbox{org}}$ & 77.70\,/\,65.30 & 64.30\,/\,46.60 & 57.90\,/\,44.30 & 45.70\,/\,29.80 & 43.80\,/\,29.70 & 57.10\,/\,38.60 &  57.80\,/\,42.40 \\
    & LAKM & 80.00\,/\,66.80 & 65.90\,/\,48.00 & 60.50\,/\,45.50 & - & - & - & - \\
    \midrule
    \multirow{3}{*}{XLM} & MLQA$_{\mbox{org}}$ & 74.90\,/\,62.40 & 68.00\,/\,49.80 & 62.20\,/\,47.60 & 54.80\,/\,36.30 & 48.80\,/\,27.30 & 61.40\,/\,41.80 &  61.70\,/\,44.20 \\
    & MLQA$_{\mbox{re-imp}}$ & 77.66\,/\,64.60 & 67.88\,/\,50.09 & 63.27\,/\,47.93 & 57.50\,/\,38.84 & 54.36\,/\,38.82 & 63.56\,/\,43.33 & 64.04\,/\,47.27 \\
    & CalibreNet (MLQA$_{\mbox{re-imp}}$) & 78.64\,/\,65.62 & 68.27\,/\,50.56 & 63.98\,/\,48.70 & 58.14\,/\,39.23 & 55.06\,/\,39.30 & 64.02\,/\,43.59 & \textbf{64.69}\,/\,\textbf{47.83}$\P$ \\
    \midrule
    \multirow{3}{*}{XLM-R$_{base}$} & MLQA$_{\mbox{org}}$  & 77.80\,/\,65.30 & 67.20\,/\,49.70 & 60.80\,/\,47.10 & 53.00\,/\,34.70 & 57.90\,/\,41.70 & 63.10\,/\,43.10 &  63.30\,/\,46.90 \\
    & MLQA$_{\mbox{re-imp}}$ & 79.21\,/\,65.96 & 67.73\,/\,49.94 & 60.92\,/\,46.31 & 55.76\,/\,36.89 & 59.58\,/\,42.92 & 66.30\,/\,45.64 & 64.92\,/\,47.94 \\
    & CalibreNet (MLQA$_{\mbox{re-imp}}$) & 79.68\,/\,66.51 & 68.04\,/\,50.77 & 61.66\,/\,47.55 & 56.14\,/\,37.83 & 59.97\,/\,43.84 & 66.92\,/\,46.59 &  \textbf{65.40}\,/\,\textbf{48.84}$\P$ \\
    \bottomrule
    \end{tabular}
    \end{center}
}
\end{table*}

\subsection{Baselines}
We use the most representative cross-lingual pre-trained language models as the initial models in our experiments, including mBERT, XLM, and XLM-R$_{base}$. Those are generic language models and thus need to go through fine-tuning before being applied to specific tasks. Sometimes, a second-round, task-oriented pre-training, such as the one in Section~\ref{chp_pbr} in our approach, may be carried out before fine-tuning. In Table~\ref{tab_main_compare}(a) and (b), the ``Model'' column refers to the generic language models and the ``Method'' column indicates the different approaches to fine-tuning or second round of pre-training. 

Coming with XGLUE-NER~\cite{Liang2020XGLUEAN} and MLQA~\cite{lewis2019mlqa}, the authors also implement their method and report the performance on the test data. We include those reported methods (\textit{i.e.}, mBERT, XLM, and XLM-R$_{base}$) as baselines in our experiments. To make the comparison fair, we also reproduce their methods using the same training data and consistent parameter settings. We report alerts if our implementation has inconsistent results to the original literature.  

To further illustrate the effect of our approach in the NER and MRC tasks, we also choose the corresponding methods that achieve state-of-the-arts results in those tasks as baselines in our experiments. By observing that CalibreNet incorporating state-of-the-art methods as the base module can further improve those methods, we show that our approach of boundary calibration complements to the ideas in the previous methods.  Specifically, we choose the following two methods.


\textbf{Single-Source KD}~\cite{wu2020single} focuses on zero-shot learning of cross-lingual NER with no labeled data in target languages.  It achieves state-of-the-art performance to the best of our knowledge. Under the teacher-student framework, the teacher model is first trained using the English labeled training set as a single source. Then, the student model is trained using the unlabeled data in the target languages with the soft label probability distribution predicted by the teacher model.

\textbf{LAKM}~\cite{yuan2020enhancing} applies unsupervised mining techniques to derive large-scale multilingual knowledge phrases (e.g., entities, phrasal N-grams, etc.) from the web. Instead of being randomly masked N-grams, those mined knowledge phrases are masked during the pre-training stage in order to improve answer boundary detection and the pre-trained model is utilized for the fine-tuning stage. Here we use the results under the zero-shot setting.

\subsection{Experiment Setup}

We implement CalibreNet by extending the code of BertForQuestionAnswering\footnote{https://github.com/huggingface/transformers}. For the AdamW~\cite{loshchilov2017decoupled} optimizer, we set the weight decay to $0.005$ for the PBR task in pre-training, to $0.0075$ for XGLUE-NER, and to $0.01$ for MLQA. The batch size is $64$ for both pre-training and fine-tuning. The initial learning rate is set to $3e-5$. We adopt the cosine learning rate scheduler with hard restarts. The warm-up proportion is set to $0.1$. When clipping the input sequence, we empirically set the maximum sequence length to $384$ for PBR and MLQA, to $192$ for XGLUE-NER. For the trade-off parameters in the loss function of the calibration module, $\alpha = \beta = 0.3$ and $\gamma = 0.4$. We train our model using 8 NVIDIA V100 GPUs with 32GB of memory and mixed precision. It takes 3 epochs on each stage for PBR pre-training and 4 epochs for fine-tuning. We save the checkpoint every 200 steps and select the best model based on the English dev set when fine-tuning CalibreNet.

\subsection{Experiment Results}

Table~\ref{tab_main_compare} shows the overall cross-lingual sequence labeling results. The ``Model'' column indicates the pre-trained language models that we build on for the sequence labeling tasks. For XGLUE-NER, we post-process the calibrated answers in each sentence to obtain the output sequence under the BIO labeling schema~\cite{ramshaw1999text} then calculate the F1 score. We report both the results from the original paper (marked with subscript ``org'') and our re-implementation (marked with subscript ``reimp''). The notation of ``CalibreNet'' with a model name in parenthesis means we use that model as the base module in CalibreNet. For example, ``CalibreNet (Sigle-source KD)'' means that the single-source KD approach~\cite{wu2020single} is employed as the base module in CarlibreNet. 

\emph{NER Results:} As shown in Table~\ref{tab_main_compare}(a), first, compared with the baselines of XGLUE-NER$_{\mbox{re-imp}}$ on top of mBERT and XLM-R$_{base}$,  CalibreNet(XGLUE-NER$_{\mbox{re-imp}}$) achieves an average F1 score gain of 1.59 and 1.92, respectively. In the languages where the Phrase Boundary Recover task is executed, that is, es and de, CalibreNet achieves significant improvements with 1.44 and 4.6 respectively.

Second, the Single-source KD model is set as the base module of CalibreNet (Single-source KD). Our model calibrates the initial answers from the distilled student models, and we search the best hyperparameter on the target language dev sets following the strategy in~\cite{wu2020single}. Our method achieves a substantial improvement of 1.13 F1 score on average comparing with Single-source KD and achieves SOTA performance on single-source cross-lingual NER.

\emph{MRC Results:} 
The results on cross-lingual MRC are shown in Table~\ref{tab_main_compare}(b). Comparing the baselines XLM and XLM-R$_{base}$, CalibreNet improves the EM score by 0.56 and 0.90 on average, respectively. For individual languages, CalibreNet (MLQA$_{\mbox{re-imp}}$) improves the EM score consistently, ranging from 0.55 for en to 1.24 for de. This clearly verifies the generalization capability of our approach to multiple languages.

In general, although the phrase boundary recover task is carried out on only en, es, and de Wikipedia corpora, CalibreNet achieves remarkable gains of the other languages in the experiments. For example, our XLM-R$_{base}$ based method has obtained improvements by 0.72 F1 score and 0.95 EM score for nl and vi. Those gains can be attributed to the multi-language encoding in the underlying mBERT, XLM, or XLM-R models. The observation that the gain on nl is larger than that on vi also matches the intuition that closely-related languages (en, de, and nl are in the same language branch) could share more knowledge within the branch, which makes transfer more efficient. 

The NER and MRC results consistently verify the effectiveness of CalibreNet for multilingual sequence labeling tasks. Besides, we conduct another analysis to study whether more gains could be obtained by applying another round of calibration. Take mBERT on XGLUE-NER as example, the average F1 further improves from 80.03 to 80.1. This shows that: (1) CalibreNet can also refine the boundaries of itself. (2) The calibration module is already quite strong at boundary detection. Thus the room for further improvement on top of itself is limited. 

Next, we further examine whether the improvement is from the capability to calibrate the initial answer boundaries, which is acquired through the CalibreNet architecture as well as the pre-training by the phrase boundary recovery task.


\section{Analysis}\label{sec:ablation}

In this section, we first discuss the objective functions and training strategies of our Phrase Boundary Recovery task.  Then, we ablate important components in CalibreNet to analyze their effects on the model. Unless otherwise specified, CalibreNet in this section is the XLM-R$_{base}$ based version on XGLUE-NER.

\subsection{Effectiveness of Pre-Training Calibration Module\label{effect_pretrain}}

\subsubsection{Phrase Boundary Recover (PBR) Task} 

\begin{table}[t]
\small
    \caption{Ablation Analysis of Pre-Training.} \label{tab_abl_pbr}
        \begin{center}
        \begin{tabular}{l|cccc|c}
        \toprule
        \textbf{Method} & \textbf{en} & \textbf{es} & \textbf{de} & \textbf{nl} & \textbf{Avg.} \\
        \midrule
        CalibreNet          & 91.00 & 78.04 & 74.10 & 80.55 & \textbf{80.92} \\
        \midrule
        \emph{- MRC} & 91.26 & 77.25 & 70.79 & 79.16 & 79.62 \\
        \emph{- MLM} & 91.09 & 76.71 & 72.25 & 78.23 & 79.57 \\
        \emph{- MRC\&MLM} & 91.25 & 77.10 & 70.00 & 79.15 & 79.38 \\
        \bottomrule
        \end{tabular}
        \end{center}
        \vspace{-8pt}
\end{table}

Some recent studies~\cite{liu2019roberta} show that simply increasing the size of pre-training data leads to performance improvement. To verify whether the gain is from data increase or the training objective, we use the same training data and the same setting to pre-train XLM-R$_{base}$ based CalibreNet without MRC (\emph{- MRC}) and Mask language modeling (\emph{- MLM}) objectives. Table~\ref{tab_abl_pbr} compares the performance on the XGLUE-NER dataset. When removing MRC and MLM objectives one at a time, the average F1 score drops by 1.30 and 1.35, respectively. It illustrates that MLM may be used as a constraint in the single MRC method, and combining MRC with MLM in our PBR task brings significant performance gain. We can observe prominent performance regression when removing the pre-training task (\emph{- MRC\&MLM}), which indicates PBR plays an important role in 
multilingual sequence labeling calibration.

\subsubsection{Continual Learning} 

Furthermore, we compare the different learning methods for the PBR task. In Table~\ref{tab_pbr_train}, sequential learning means training a model on a monolingual corpus sequentially, and ``mixed'' means we mix all the corpora to train a model. Compared with sequential learning, continual learning keeps part of the current monolingual training corpus for the following stages to avoid catastrophic forgetting and confusion. The F1 score of mixed learning on en is 0.54 higher than continual learning, while our method generally performs better on target languages (low resource).

\begin{table}[t]
\small
    \caption{Comparison of PBR with different training strategies.} \label{tab_pbr_train}
    \begin{center}
        \begin{tabular}{l|cccc|c}
        \toprule
        \textbf{Strategy} & \textbf{en} & \textbf{es} & \textbf{de} & \textbf{nl} & \textbf{Avg.} \\
        \midrule
        Continual learning  & 91.00 & 78.04 & 74.10 & 80.55 & \textbf{80.92} \\
        \midrule
        Sequential learning & 90.46 & 77.64 & 74.35 & 78.47 & 80.23 \\
        Mixed learning      & 91.54 & 77.20 & 73.93 & 80.28 & 80.74 \\
        \bottomrule
        \end{tabular}
    \end{center}
\end{table}

\begin{table}[h]
\small
    \caption{Ablation Analysis of CalibreNet on XGLUE-NER.  - denotes removing the corresponding component from CalibreNet.} \label{tab_abl_calinet}
    \begin{center}
    \begin{tabular}{l|cccc|c}
    \toprule
    \textbf{Methods} & \textbf{en} & \textbf{es} & \textbf{de} & \textbf{nl} & \textbf{Avg.} \\
    \midrule
    CalibreNet & 91.00 & 78.04 & 74.10 & 80.55 & \textbf{80.92} \\
    \midrule
    \emph{- Index}                & 90.58 & 77.63 & 71.56 & 80.11 & 80.04 \\
    \emph{- CLS}                  & 91.44 & 77.44 & 70.61 & 78.93 & 79.61 \\
    \emph{- Matching}             & 91.89 & 77.36 & 73.52 & 79.89 & 80.67 \\
    \emph{- Index\&Matching}      & 90.58 & 77.62 & 71.21 & 78.95 & 79.59 \\
    \emph{- Index\&CLS}           & 90.70 & 76.20 & 70.24 & 78.64 & 78.95 \\
    \emph{- Index\&Matching\&CLS} & 90.83 & 76.22 & 70.15 & 78.64 & 78.96 \\
    \bottomrule
    \end{tabular}
    \end{center}
    \vspace{-5pt}
\end{table}

\begin{table}[h]
\small
    \caption{Comparison of CalibreNet with ensemble baseline method on XGLUE-NER. * denotes XLM-R$_{base}$.} \label{tab_comp_ensemble}
    \resizebox{1.0\linewidth}{!}{
    \begin{tabular}{ll|cccc|c}
    \toprule
    \textbf{Model} & \textbf{Methods} & \textbf{en} & \textbf{es} & \textbf{de} & \textbf{nl} & \textbf{Avg.} \\
    \midrule
    \multirow{4}{*}{mBERT}  & XGLUE-NER$_{\mbox{re-imp}}$\textsubscript{1} & 90.59 & 74.84 & 70.46 & 77.87 & 78.44 \\
    & XGLUE-NER$_{\mbox{re-imp}}$\textsubscript{2} & 90.88 & 74.91 & 69.79 & 78.00 & 78.40\\
    & Ensemble   & 91.05 & 75.91 & 71.33 & 78.89 & 79.29 \\
    & CalibreNet & 91.25 & 76.10 & 73.06 & 79.70 & \textbf{80.03} \\
    \midrule
    \multirow{4}{*}{XLM-R*} & XGLUE-NER$_{\mbox{re-imp}}$\textsubscript{1} & 91.20 & 76.60 & 69.50 & 78.69 & 79.00 \\
    & XGLUE-NER$_{\mbox{re-imp}}$\textsubscript{2}   & 91.18 & 76.65 & 69.74 & 78.83 & 79.10 \\
    & Ensemble   & 91.61 & 77.09 & 70.65 & 78.78 & 79.53 \\
    & CalibreNet & 91.00 & 78.04 & 74.10 & 80.55 & \textbf{80.92} \\
    \bottomrule
    \end{tabular}
    }
\end{table}

\subsection{Ablation Analysis of CalibreNet Architecture}
To unveil the contribution of each component, we conduct an ablation study, whose results are reported in Table~\ref{tab_abl_calinet}. Each of initial-index embedding (\emph{-Index}), entity type classification (\emph{-CLS}), and span matching (\emph{- Matching}) makes an important contribution to the overall performance. We observe that span matching (\emph{- Index\&CLS}) does not perform as well as we expect when it is used alone. However, combining span matching with other components still improves the average F1 score when comparing \emph{- Index\&Matching} and \emph{- Index}. Especially, after removing all additional components (\emph{- Index\&Matching\&CLS}) the performance of vanilla MRC drops below the baseline model, which fully illustrates the problems mentioned in Section~\ref{chp_archi}.

\subsection{Comparison with Ensemble Methods}
In the last part of analysis, we investigate the impact of model ensemble. First, we use the same hyperparameters and different random seeds to train two baseline models:  XGLUE-NER$_{\mbox{re-imp}}$\textsubscript{1} and XGLUE-NER$_{\mbox{re-imp}}$\textsubscript{2}. Second, the predicted distributions of the two models are averaged to obtain the Ensemble model. As shown in Table~\ref{tab_comp_ensemble}, the ensemble method performs much worse compared with our method, which suggests the gain of CalibreNet is not from the simple ensemble of two models. Instead, the calibration procedure is effective in improving the overall performance.

\section{Conclusion}\label{sec:conclusion}

In this paper, we tackle the challenge of detecting span boundaries more precisely for sequence labeling tasks in low-resource languages. We propose the CalibreNet architecture as well as a novel Phrase Boundary Recovery task for more accurate boundary detection. Extensive experimental results verify the effectiveness of our approach and the generalization capability for multiple languages. As future works, we plan to introduce entity type prediction in the pre-training task, and also develop better methods for question generation for the MRC task.

\begin{acks}
Jian Pei's research is supported in part by the NSERC Discovery Grant program. Shining Liang's research is partially sponsored by the National Natural Science Foundation of China (61976103, 61872161), the Scientific and Technological Development Program of Jilin Province (20190302029GX, 20180101330JC, 20180101328JC). All opinions, findings, conclusions and recommendations in this paper are those of the authors and do not necessarily reflect the views of the funding agencies.
\end{acks}

\bibliographystyle{ACM-Reference-Format}
\bibliography{sample-base}









\end{sloppy}
\end{document}